\documentclass[preprint,12pt]{elsarticle}

\usepackage{amssymb}

\usepackage{amsfonts}       
\usepackage{nicefrac}       
\usepackage{microtype}      
\usepackage{xcolor}         
\usepackage{amsmath}
\usepackage{dsfont}
\usepackage{graphicx}
\usepackage{multirow}
\usepackage{subcaption}
\usepackage[ruled,vlined]{algorithm2e}
\usepackage{setspace}
\usepackage{tikz}
\usetikzlibrary{mindmap}

\usepackage{pifont}
\newcommand{\cmark}{\ding{51}}%
\newcommand{\xmark}{\ding{55}}%

\journal{Pattern Recognition}

\begin{document}

\begin{frontmatter}



\title{Going Deeper into Semi-supervised Person Re-identification}

\author[qut]{Olga Moskvyak\corref{c1}}
\cortext[c1]{Corresponding author}
\ead{olga.moskvyak@hdr.qut.edu.edu}
\author[qut]{Frederic Maire}
\author[qut]{Feras Dayoub}
\author[uq]{Mahsa Baktashmotlagh}

\address[qut]{School of Electrical Engineering and Robotics, Queensland University of Technology, \\2 George Street, Brisbane City, QLD 4000, Australia}
\address[uq]{School of Information Technology and Electrical Engineering, The University of Queensland, St Lucia, QLD 4072, Australia}




\begin{abstract}
Person re-identification is the challenging task of identifying a person across different camera views.
Training a convolutional neural network (CNN) for this task requires annotating a large dataset, and hence, it involves the time-consuming manual matching of people across cameras. To reduce the need for labeled data, we focus on a semi-supervised approach that requires only a subset of the training data to be labeled. We conduct a comprehensive survey in the area of person re-identification with limited labels. Existing works in this realm are limited in the sense that they utilize features from multiple CNNs and require the number of identities in the unlabeled data to be known. To overcome these limitations, we propose to employ part-based features from a single CNN without requiring the knowledge of the label space (i.e., the number of identities). This makes our approach more suitable for practical scenarios, and it significantly reduces the need for computational resources.
We also propose a PartMixUp loss that improves the discriminative ability of learned part-based features for pseudo-labeling in semi-supervised settings.
Our method outperforms the state-of-the-art results on three large-scale person re-id datasets and achieves the same level of performance as fully supervised methods with only one-third of labeled identities.

\end{abstract}




\end{frontmatter}

\section{Introduction}

The person re-identification (re-id) task aims at matching images of the same person captured by non-overlapping surveillance cameras.
Convolutional neural networks (CNN) have demonstrated good results on large-scale cross-camera annotated datasets of pedestrians \citep{market-dataset, duke-dataset, part-loss, Zhang2019DenselySA, batch-hard-triplet, bag-tricks-person-reid, Ghorbel2020FusingLA}.
However, annotating person identities in multiple camera views is a labor-intensive task in practical scenarios. Therefore, a more realistic setting called \textit{semi-supervised person re-identification} has gained attention in recent years, where a subset of data is annotated across cameras, and the rest of the data is used without labels.

\textit{Pseudo-labeling}\footnote{Assigning labels to unlabeled data is called \textit{pseudo-labeling}, and the computed labels are called \textit{pseudo-labels}.} is one of the key components in the recent semi-supervised learning approaches~\cite{Iscen2019LabelPF}. However, a critical challenge of pseudo-labeling for semi-supervised person re-id is that the number of identities is unknown in the unlabeled subset. 
Existing works aim to solve this problem by fixing the number of unlabeled identities during training \citep{SSL-person-reid-multi-view, SSL-Xin2019DeepSL}, or by utilizing features from multiple CNNs to improve the quality of pseudo-labels \citep{SSL-person-reid-multi-view, SSL-Xin2019DeepSL, SSL-Pan2019MultiViewAM}. These approaches are limited in the sense that making an assumption on the number of unlabeled identities is not feasible in practical scenarios. Moreover, ensembling multiple networks is computationally intensive, memory-wise and time-wise, at the training and test stage.

To overcome these limitations, we propose a pseudo-labeling method that does not require the number of unlabeled identities to be known/assumed. In contrast to network ensembling approaches, we propose to use a single model, which significantly reduces the space and time complexity. 
Motivated by the success of part-based features for supervised learning \citep{Sun2018BeyondPM, Zhang2019DenselySA, part-loss, pcb-model}, we propose to employ part-based embeddings for pseudo-labeling in semi-supervised settings. 
We consider semantic parts of the image, and a pseudo-label is derived based on consensus clustering of parts embeddings. 
Our method does not require a sophisticated part detector and performs well with coarse part detection such as equal horizontal stripes \citep{pcb-model} because images of people are mostly walking pedestrians.
To the best of our knowledge, our method is the first to utilize embeddings of semantic parts to compute pseudo-labels.

Representations for person re-id are commonly learned with a triplet loss \cite{triplet-loss}, its variations \cite{batch-hard-triplet, Chen2017BeyondTL} and combinations with cross-entropy loss \cite{bag-tricks-person-reid, Zhai2019InDO}.
The key challenge in learning embeddings with the triplet loss is mining hard triplets that contribute non-zero value to the loss \cite{batch-hard-triplet}.
To improve part-based embeddings' learning and discriminative ability, we propose a \textit{PartMixUp loss} function that minimizes the distances between more difficult training samples compared to the triplet loss.
We utilize the observation that for the pair of pedestrian images to be negative (images from different people), it is sufficient that only one semantic part is different.
For example, if only the shoes are different, it is considered a different identity in person re-identification.
Therefore, we create training pairs that contribute non-zero value by sampling an image of a person and its copy where some semantic parts are replaced with corresponding parts from a different random identity.
We perform this operation on the embedding level and do not manipulate input images.
The advantage of this technique is that it can be used on both labeled and unlabeled images, provided that identities are disjoint to avoid replacing a part from the same identity.

The main assumptions that we make in this work are as follows:
(1) the labeled and unlabeled data comes from the same domain. This represents a real-world use case when only a subset of the collected data is labeled;
(2) the labeled and unlabeled subsets have disjoint identities, representing a practical scenario where annotated and unlabeled images come from different time periods.

In summary, the main contributions of the paper are three-fold: 
\begin{itemize}
    \item we are the first to conduct an in-depth survey of existing works in person re-id with limited annotations and identify their advantages and limitations;
    \item we propose a novel method for semi-supervised person re-identification based on consensus clustering of embeddings for semantic parts that do not make an assumption about the number of identities in the unlabeled subset; 
    \item we introduce a PartMixUp loss for enhancing the learning of discriminative embeddings by mixing up embeddings of semantic parts.
\end{itemize}

The rest of the paper is organized as follows.
We first conduct a comprehensive overview of person re-id from the perspective of the amount of annotated data in Section \ref{sec:related-work}.
Then we describe our proposed semi-supervised training method for the re-id task and PartMixUp loss in Section \ref{sec:methodology}.
Finally, we demonstrate the results of our experiments, evaluate the method's components and discuss ablation studies in Section \ref{sec:experiments}.

\section{Related work}
\label{sec:related-work}

\begin{figure}[!h]
\centering
\resizebox{0.85\textwidth}{!}{
\begin{tikzpicture}[mindmap, grow cyclic, every node/.style=concept, concept color=orange!40, 
	level 1/.append style={level distance=4.3cm,sibling angle=90},
	level 2/.append style={level distance=3cm,sibling angle=45},]

\node{Limited-labels re-id}
child { node {Features}
	child [grow = -45] { node {Multi-view \citep{SSL-person-reid-multi-view, SSL-Xin2019DeepSL, SSL-Pan2019MultiViewAM}}}
	child [concept color=green!40, grow = -90] { node {Part-based (Ours)}}
}
child { node {Extra data}
	child [grow = 45] { node {GAN-based \cite{Ding2019FeatureAP, Zhang2020SemisupervisedPR}}}
	child [grow = 90] { node {Domain adaptation \cite{Ge2020SelfpacedCL}}}
}
child { node {Pseudo-labeling}
	child { node {K-means \citep{Fan2018UnsupervisedPR, SSL-person-reid-multi-view, SSL-Xin2019DeepSL}}}
	child { node {KNN graph \cite{SSL-Pan2019MultiViewAM, SSL-Chang2020TransductiveSM}}}
	child { node {Affinity Propagation \citep{SSL-Liu2020SemanticsGuidedCW}}}
	child [concept color=green!40] { node {Agglome-\\rative (Ours)} }
}
child  { node {Level of supervision}
	child { node {Intra-camera labels \cite{Qi2020ProgressiveCS, Wang2021GraphInducedCL, Zhu2019IntraCameraSP}}}
	child { node {One-shot per person \cite{Li2018SemisupervisedRM, Bk2017OneShotML}}}
	child [concept color=green!40] { node {Semi-supervised \citep{SSL-Chang2020TransductiveSM, SSL-Pan2019MultiViewAM, SSL-person-reid-multi-view}, Ours}}
	child { node {Unsuper-\\vised \citep{Fan2018UnsupervisedPR, Lin2020UnsupervisedPR}}}
};
\end{tikzpicture}
}
\caption{Limited labels re-id from different perspectives. We highlight the place of our work in respect of existing works (in green).}
\label{fig:related-work-mind-map}
\end{figure}
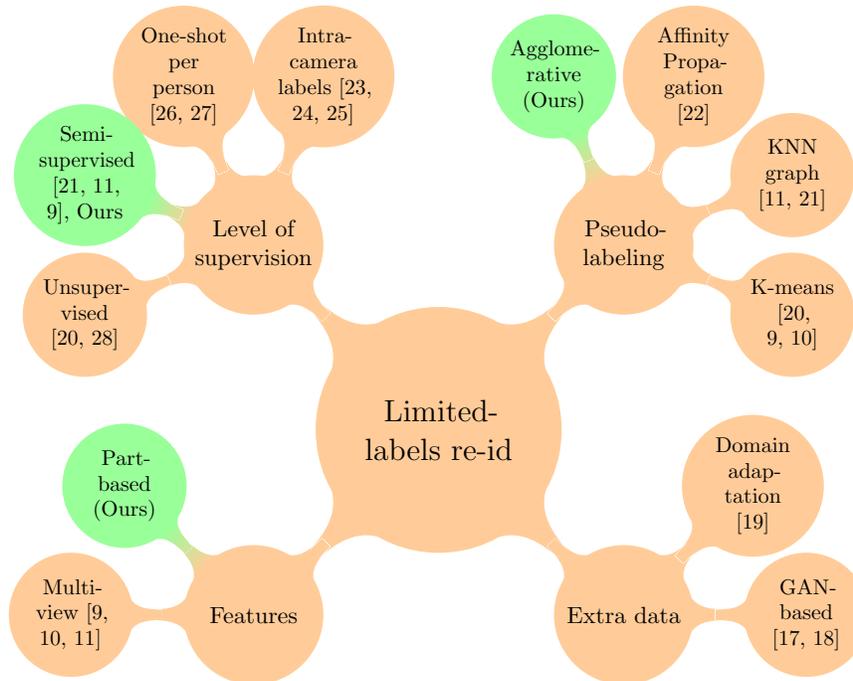

Person re-identification is a wide research area that has been studied from various perspectives including deep feature learning, ranking optimization and metric learning.
Recent surveys \citep{Ye2021DeepLF, Leng2020ASO, Yaghoubi2021SSSPRAS} have conducted a comprehensive overview and systematize existing works in person re-id. 
Different from previous surveys, we specifically focus on the limited labels scenario which is a less studied research.
We highlight the assumptions and limitations of previous methods that constraint their applications to practical scenarios and propose a method that overcomes the limitations of existing works.
We review person re-id methods from various perspectives including: the level of supervision (e.g., amount of labeled data), pseudo-labeling strategies, feature learning process (i.e., features learnt from multiple or single networks), and if they generate extra data for training or not.
Finally, we emphasize the place of our work with respect to existing works in Figure~\ref{fig:related-work-mind-map}.

\subsection{Level of supervision}

Due to the performance plateau of fully supervised methods for person re-id, the focus of the recent research has shifted to purely unsupervised, semi-supervised, one-shot per person, and intra-camera labels settings.

Unsupervised person re-id \citep{Fan2018UnsupervisedPR} is a challenging task due to the absence of target labels.
Existing works utilize clustering to learn pseudo-labels \citep{Lin2019ABC}, introduce learning soft similarity labels \citep{Lin2020UnsupervisedPR} and employ deep asymmetric metric learning \citep{Yu2020UnsupervisedPR}.
However, purely unsupervised methods show inferior performance to methods with any amount of annotations.

Intra-camera labels (i.e., images of the person within one camera) are an attractive scenario as intra-camera labels are easy to obtain using tracking algorithms \cite{Maksai2017NonMarkovianGC} while cross-camera labels require human effort.
However, a model trained only on intra-camera labels tends to learn camera-specific features and fails to generalize across cameras \cite{Qi2020ProgressiveCS}.
To learn camera agnostic features, Qi et al. \cite{Qi2020ProgressiveCS} propose progressive learning of cross-camera soft labels and
Zhu et al. \cite{Zhu2019IntraCameraSP} design a method targeted for self-discovering the inter-camera identity correspondence.
Another scenario is labeling one example per identity \cite{Li2018SemisupervisedRM, Bk2017OneShotML} assuming that a few images per identity is labeled and the rest of the training images are without labels.

Although the aforementioned scenarios eliminate the tedious inter-camera identity labeling process, they require at least one example per identity to be present in the labeled training subset.
Thus, we focus our attention on a semi-supervised scenario where a small subset (e.g., 10-30\%) of data is annotated with cross-camera and within-camera labels.

\subsection{Pseudo-labeling strategies}

Pseudo-labeling is the process of assigning labels to unlabeled examples, which has been successfully applied in the classification tasks where there are a fixed list of class labels.
Pseudo-labeling for person re-id is challenging as there exists an unknown number of identities in the unlabeled subset.
Existing works utilize either k-means clustering \citep{Fan2018UnsupervisedPR, SSL-person-reid-multi-view, SSL-Xin2019DeepSL} or kNN graphs \cite{SSL-Pan2019MultiViewAM, SSL-Chang2020TransductiveSM} which work under the assumption that the number of clusters (i.e., the number of identities) is known in the unlabeled data.
Unsurprisingly, the best performance is achieved where the number of clusters equals the number of true identities in the unlabeled data. 
Although the methods are robust to some variations in the number of assumed clusters, this requirement significantly limits their applicability in practical scenarios.
In this work, we remove the requirement of prior knowledge about the number of clusters in unlabeled data.

\subsection{Feature learning process}

Several works utilize ensembles of neural networks to learn multi-view features and obtain pseudo-labels for the unlabeled subset \citep{SSL-person-reid-multi-view, SSL-Xin2019DeepSL, SSL-Pan2019MultiViewAM}.
However, training multiple CNNs increases the usage of memory and computational resources.
Moreover, ensembling methods are usually superior to single models, and as such, it is not clear how much of the performance gain comes from the method itself as opposed to using feature ensembling.
Different from previous works, our approach learns embeddings and computes pseudo-labels from training a single CNN.
Inspired by successful usage of part-based features \cite{part-loss, pcb-model, Ghorbel2020FusingLA, Eom2019LearningDR} in supervised person re-id, we employ part embeddings for assigning pseudo-labels in semi-supervised settings.

\subsection{Generative frameworks}

A compelling approach is to generate additional labeled data using the Generative Adversarial Networks (GANs) \citep{Zheng2017UnlabeledSG}: Ding et al. \citep{Ding2019FeatureAP} considers feature affinities between GAN's generated samples and labeled data to estimate labels; and Zheng et al. \citep{Zheng2019JointDA} proposes an end-to-end joint learning framework for training re-id and data generation tasks.
However, challenges in generating images that depict the same identity in different poses 
limit the applicability of this method.

Another way to overcome the problem of scarcity of labeled data is employing a deep re-id model pre-trained on a labeled domain and transferring the knowledge to the label-scarce domain by reducing the domain discrepancy between the two domains~\cite{Deng2018ImageImageDA}.
While the aforementioned unsupervised domain adaptation strategy yields impressive performance \citep{Ge2020SelfpacedCL, Yu2019UnsupervisedPR, Zhong2019InvarianceME}, fully annotated large datasets with similar identities may not be available in many practical scenarios.
Therefore, we focus on semi-supervised learning from only one domain to alleviate the need for an external dataset or identity annotation.
    
\section{Methodology}
\label{sec:methodology}

\subsection{Pseudo-labeling via consensus clustering of semantic part embeddings}
\label{sec:methodology-pseudo-labeling}

\begin{figure*}[t]
    \begin{center}
       \includegraphics[width=0.99\linewidth]{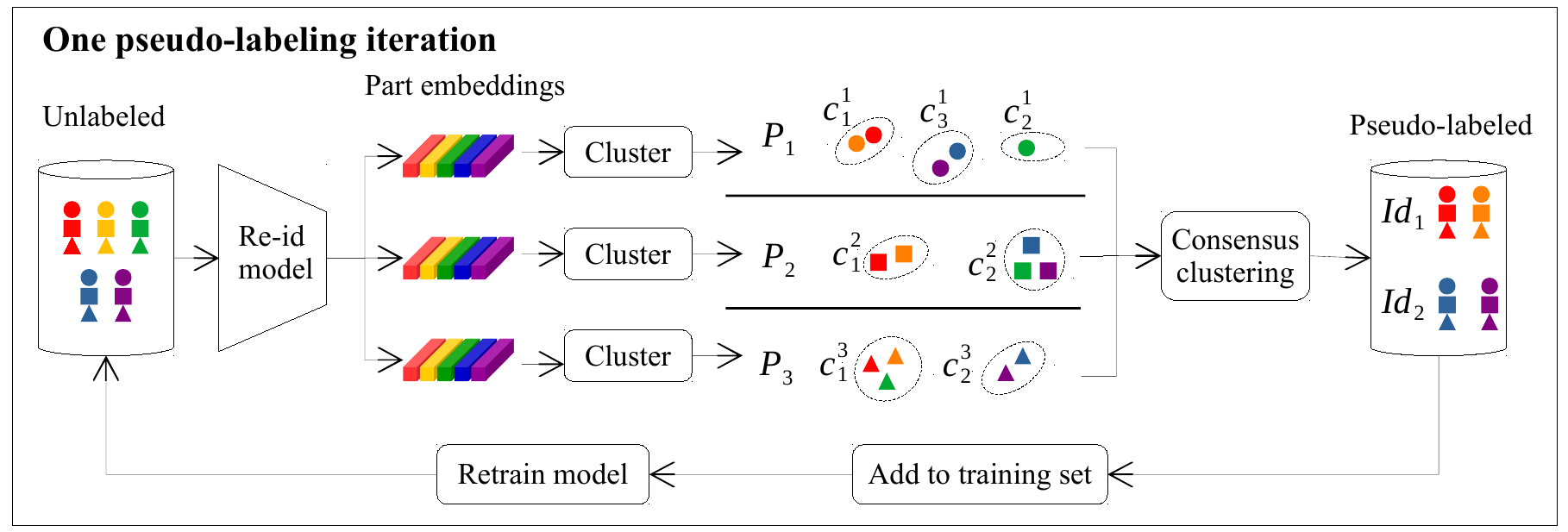}
    \end{center}
       \caption{One pseudo-labeling iteration via consensus clustering of part-based embeddings. Each color represents an unknown identity in the unlabeled subset. A circle, a square and a triangle represent semantic parts of the image. The model is re-trained on the union of the labeled (not shown in the figure) and the pseudo-labeled subsets.
       }
    \label{fig:part-based-clustering}
\end{figure*}

We employ part-level features and propose a \textit{consensus clustering of semantic part embeddings} for pseudo-labeling in semi-supervised person re-identification.
By clustering embeddings for semantic image parts, each image gets a list of cluster assignments $(c^1, c^2, \dots, c^Q)$ where $Q$ denotes the number of semantic parts.
The list $(c^1, c^2, \dots, c^Q)$ can be seen as an encoded part description of a person.
For example, if we consider a coarse partition in three body parts (head, upper body, legs), the cluster assignment can be interpreted as (dark hair, white top, black bottom).
Embeddings of semantic parts are clustered independently, and pseudo-labels for images are determined based on the agreement between parts' clusters.

Our method can be used with any convolutional model that outputs part-based embeddings such as PCB \citep{pcb-model}, DPB\citep{part-loss} or KAE-Net \citep{KAE-Net}.
The output of a model compatible with our method should be an array of part embeddings $[h^1, h^2, \dots, h^Q]$, with $h^{i} \in \mathbb{R}^{d}$, $Q$ the number of parts, and $d$ the dimension of the embedding space for semantic parts.

The model is optimized in multiple \textit{pseudo-labeling iterations}.
In particular, the model is retrained at each iteration using both labeled data and a subset of unlabeled data with computed pseudo-labels.
The following section explains training steps in one pseudo-labeling iteration.

\subsubsection{Training steps in one pseudo-labeling iteration}
\label{sec:methodology-training-schedule}

At each pseudo-labeling iteration, a round of model training, clustering embeddings, and assigning pseudo-labels to the unlabeled subset is performed, as illustrated in Figure~\ref{fig:part-based-clustering}.
Let $(X^{\text{L}}, Y^{\text{L}})$ be the labeled data with corresponding labels and $X^{\text{U}}$ be the unlabeled data.
$(X^{\text{PL}}, Y^{\text{PL}})$ denotes the pseudo-labeled subset which is empty at the start of the algorithm.

At the beginning of each pseudo-labeling iteration, the model $f_{\theta}(x)$ is initialized with ImageNet~\citep{imagenet} pre-trained weights.
We found experimentally that re-initializing the model at the start of each pseudo-labeling iteration yields better performance than fine-tuning from the previous iteration.
The model is optimized on the union of labeled subset $(X^{\text{L}}, Y^{\text{L}})$ and pseudo-labeled subset $(X^{\text{PL}}, Y^{\text{PL}})$ by minimizing the loss of Eq.~\eqref{eq:total-loss}.
Once the model $f_{\theta}$ has converged, we compute part embeddings for unlabeled images
$h^q_i = f_{\theta}(x_i)$, with $x_i \in X^{\text{U}}$ and $q \in (1, \dots Q)$.

Embeddings $[h^q_1, h^q_2, \dots, h^q_{|X^{\text{U}}|}]$ for each semantic part $q$ are clustered independently. 
We explain the clustering step in detail in Section~\ref{sec:methodology-clustering}.
The clustering result is a list of partitions $[P_1, P_2, \dots, P_Q]$.
Each partition $P_q = [c_1^q, c_2^q, \dots, c_{k_q}^q]$ consists of $k_q$ clusters so that the cluster assignment is disjoint $c_i^q \cap c_j^q = \emptyset$ and covers the whole unlabeled subset $\cup_{j=1}^{k_q} c_j^q = X^{\text{U}}$.
Note that the number of clusters $k_q$ for each semantic part is different and determined during clustering.

The final step in the pseudo-labeling iteration is to aggregate partitions for each semantic part to obtain image-level pseudo-labels.
We use consensus clustering \citep{Alqurashi2017ClusteringEM} to aggregate multiple clustering results on a list of partitions $[P_1, P_2, \dots, P_Q]$. 
Consensus clustering aims to find a partition $P^{*}$ of the unlabeled subset $X^{\text{U}}$ by combining ensemble members $[P_1, P_2, \dots, P_Q]$ so that $P^{*}$ produces better pseudo-labels than each individual partition $P_j$. Details of consensus clustering are covered in Section~\ref{sec:methodology-consensus-clustering}.
We then obtain pseudo-labels from computed consensus clusters.
A pseudo-labeled subset $(X^{\text{PL}}, Y^{\text{PL}})$ is re-initialized with samples that have a sufficient number of examples per pseudo-label (e.g., five images per identity). Implementation details are given in Section~\ref{sec:experiments-implementation}).

After assigning samples to a pseudo-labeling subset, we proceed with the next pseudo-labeling iteration.
The pseudo-code for the whole algorithm is presented in Algorithm~\ref{alg:ss-reid}.
In the following sections, we review the clustering algorithm employed to cluster part embeddings. We then describe consensus clustering to aggregate part partitions.

\begin{algorithm}[!h]
\SetAlgoLined
\SetKwInput{KwInput}{Require}               
\SetKwInput{KwOutput}{Return}             
\SetKwBlock{Repeat}{Repeat}{until}

\KwInput{$(X^{\text{L}}, Y^{\text{L}})$ - a set of labeled images with labels}
\KwInput{$X^{\text{U}}$ - a set of unlabeled images}
\KwInput{$f_{\theta}$ - a neural network with trainable parameters $\theta$}
\KwInput{$l$ - a minimum number of members in a cluster to be considered as a pseudo-label}

$(X^{\text{PL}}, Y^{\text{PL}}) = (\emptyset, \emptyset)$

\Repeat
{ 
    Initialize $f_{\theta}$ with ImageNet pretrained weights;
    
    \For{e in [1, num\_epochs]}
    {
        Optimize $f_{\theta}$ on $(X^{\text{L}}, Y^{\text{L}}) \cup (X^{\text{PL}}, Y^{\text{PL}})$ with the loss in Equation~(\ref{eq:total-loss})
    }

    Compute embeddings on the unlabeled subset $\{h^q_1, h^q_2, \dots, h^q_{|X^{\text{U}}|}\} = f_{\theta}(X^{\text{U}})$

    Cluster embeddings $\{h^q_1, h^q_2, \dots, h^q_{|X^{\text{U}}|}\}$ to get partitions $P_q$ for all $q \in (1,\dots, Q)$

    Compute $\text{CA}$ matrix on partitions $[P_1, P_2, \dots, P_Q]$ with Equation~(\ref{eq:ca-matrix})
    
    Cluster matrix $\mathds{1}-\text{M}$ to get a partition $P^{*}$

    Re-init a pseudo-labeled subset $(X^{\text{PL}}, Y^{\text{PL}}) = (\emptyset, \emptyset)$
    
    Assign $(X^{\text{PL}}, Y^{\text{PL}}) \leftarrow (x, P^{*}(x))$ for all $x \in X^{\text{U}}$ \text{if the number of images in the cluster} $P^{*}(x)$ \text{is greater or equal to} $l$
    
} (convergence or maximum iterations are reached)

\KwOutput{$f_{\theta}$}

 \caption{Semi-supervised training of a CNN for person re-id via consensus clustering of semantic part embeddings}
 \label{alg:ss-reid}
\end{algorithm}

\subsubsection{Clustering part embeddings}
\label{sec:methodology-clustering}

We cluster embeddings for each semantic part independently using hierarchical  Agglomerative (``bottom-up") \citep{szekely2005hierarchical} clustering algorithm. Each observation starts in its own cluster, and pairs of clusters are merged as one moves up the hierarchy.
The clusters are linked using Ward’s minimum variance method \cite{ward1963hierarchical} that minimizes the total within-cluster variance.
Agglomerative clustering doesn't require the predefined number of clusters. As a criteria to merge clusters, we provide maximum distance of the clusters which we empirically set to 2~\footnote{The maximum distance between any pair is 2 because part embeddings are normalized to have the length of one.} in the experiments
The resulting number of clusters is different for each semantic part.

Apart from the agglomerative clustering, we analyze the suitability of Affinity Propagation \citep{AP} and DBSCAN \citep{DBSCAN} clustering algorithms.
Previous work \citep{Ding2019FeatureAP} uses Affinity Propagation with tuned preference values for each data point heuristically.
We favor Agglomerative clustering over Affinity Propagation and DBSCAN as it has fewer hyperparameters that require tuning using heuristics.
Ablation studies for other clustering algorithms are summarized in Section~\ref{sec:experiments-ablation}.

\subsubsection{Consensus clustering}
\label{sec:methodology-consensus-clustering}

Consensus clustering of semantic part partitions $[P_1, P_2, \dots, P_Q]$ is based on a co-association method \citep{Alqurashi2017ClusteringEM} that is recommended when the number of clusters in each partition is different \citep{VegaPons2011ASO}.
Co-association matrix $\text{M}$ counts the number of partitions when $x_i$ and $x_j$ are in the same cluster. Matrix $M$ is obtained from:
\begin{equation}
\label{eq:ca-matrix}
    \text{M}_{ij} = \frac{1}{P} \sum_{q=1}^{P} \delta(P_q(x_i), P_q(x_j))
\end{equation}
with $P_q(x_i)$ representing the associated cluster of the image $x_i$ in the partition $P_q$, and $\delta(a,b) = 1$ if $a = b$, and $0$ otherwise.

Each value in $\text{M}$ matrix is a measure of how many semantic parts of images $x_i$ and $x_j$ are in the same cluster.
Matrix $\mathds{1}-\text{M}$ can be considered as a new measure between images with values ranging from 0 (similar) to 1 (different).
The consensus partition is obtained by applying a hierarchical clustering algorithm \citep{Fred2005CombiningMC} to $\mathds{1}-\text{M}$ matrix and varying a distance threshold when two clusters can be merged.
We vary a threshold from \textit{strict} (any value below $1/Q$), meaning that all semantic parts are required to agree on the cluster assignment to \textit{intermediate} value, when a majority (e.g., 75\% of semantic parts) agree on the assignment.

The advantage of our method over previous work \citep{SSL-Liu2020SemanticsGuidedCW} is that it does not transfer any bias from the labeled subset because our pseudo-labeling method works solely on the unlabeled subset.

\subsection{PartMixUp loss function}
\label{sec:methodology-partmixup-loss}

In order to increase the discriminative ability of part embeddings, we introduce a  \textit{PartMixUp loss} that extends the triplet loss \cite{triplet-loss} for part embeddings.
We briefly review the triplet loss and outline its shortcomings, followed by our proposed PartMixUp loss that addresses the specified drawbacks.

\textbf{Triplet loss.}
The triplet loss \citep{triplet-loss} accepts triplets of images $(x_a, x_{pos}, x_{neg})$ where $x_a$ (anchor) and $x_{pos}$ (positive) are images from the same person and an image $x_{neg}$ (negative) is from a different person. 
The triplet loss $\mathcal{L}_{\text{T}}$ encourages the distances between positive pairs of embeddings to become smaller than the distances between negative pairs of embeddings by a given margin $m$:
\begin{equation} \label{eq:triplet-loss}
    \mathcal{L}_{\text{T}}(x_a, x_{pos}, x_{neg}) = 
    \max \Big(0, 
              m + 
              D\big(f(x_a), f(x_{pos})\big) - 
              D\big(f(x_a), f(x_{neg})\big)
          \Big)
\end{equation}
where $D$ is a distance metric in the embedding space (e.g., Euclidean or cosine) and $f$ is a model. 
The squared distance is commonly used to simplify the derivative computations during backpropagation.

The strategy for selecting triplets $(x_a, x_{pos}, x_{neg})$ for the triplet loss is important. 
Generating random triplets would result in many triplets already in a correct position (a negative sample is further than a positive from the anchor by a margin $m$) and contribute zero value to Equation~(\ref{eq:triplet-loss}).
Batch-hard triplet mining \citep{batch-hard-triplet} aims to overcome this problem and selects the hardest positive (the furthest example from the same class) and the hardest negative (the closest example from a different class) within a batch for each anchor image.

\textbf{PartMixUp loss.}
Our proposed PartMixUp loss $\mathcal{L}_{\text{PM}}$ aims to further improve the triplet loss by taking advantage of semantic part embeddings.
Learning discriminative part embeddings is essential for our part-based clustering, where each part contributes to the identity assignment.

\begin{figure*}[t]
    \begin{center}
       \includegraphics[width=\linewidth]{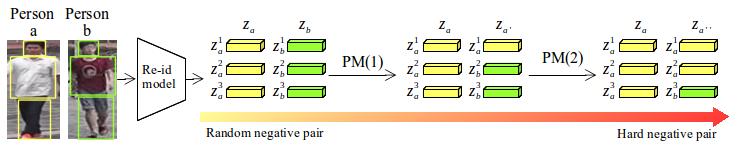}
    \end{center}
       \caption{PartMixUp (PM) loss mixes part embeddings to generate hard negative pairs that would contribute non-zero value to the loss. The number of shared parts controls the difficulty of negative pairs. Parts are replaced on the embedding level and not on the input pixel level. Part's bounding boxes are for illustrative purposes only. Three parts are shown to simplify the visualization. Our method uses six equal horizontal stripes as a coarse part detection.
       }
    \label{fig:partmixup}
\end{figure*}

PartMixUp loss builds on the observation that two different persons with similar appearances are hard to distinguish and represent useful examples for a learning algorithm.
However, it is hard to mine such pairs from the dataset.
We take advantage of part-based embeddings and generate such pairs for the PartMixUp loss by replacing some part embeddings of the image with the corresponding part embeddings from another person (Figure~\ref{fig:partmixup}). 
The created example corresponds to a new person that differs from the original only by replaced parts.

We formally demonstrate that PartMixUp loss mines hard pairs that contribute non-zero values to the loss.
Let us consider a pair of part-based embeddings from different identities $z_a = \{z^1_a, z^2_a, \dots, z^Q_a\}$ and \\ $z_{neg} = \{z^1_{neg}, z^2_{neg}, \dots, z^Q_{neg}\}$.
Semantic parts selected for replacement can be represented as two subsets $U^{\prime}$ and $U^{\prime\prime}$ of a set of indices $U = (1, 2, \dots, Q)$ so that 
$U^{\prime} \subseteq U^{\prime\prime} \subseteq U$.
The part-based embeddings $z_{a^{\prime}}$ and $z_{a^{\prime\prime}}$ are created from $z_a$ by replacing with $z_{neg}$ parts with indices in $U^{\prime}$ and $U^{\prime\prime}$, respectively.
More specifically, embeddings $z_a$ and $z_{a^{\prime}}$ have the same part embeddings for part indices in $U^{\prime}$ and different for other parts.
The same part embeddings contribute zero to the sum so the distance between a negative pair decreases when the number of similar parts increases:
\begin{equation}
\label{eq:partmixup-reduce-distance}
    D(z_a, z_{neg}) \geq D(z_a, z_{a^{\prime}}) \geq D(z_a, z_{a^{\prime\prime}})
\end{equation}

In other words, the more semantic parts are shared between images of two people, the harder it is to distinguish these people.
For example, if two individuals are dressed the same and the only difference is in the face and hairstyle, then it is a hard pair to distinguish.

The formula for computing PartMixUp loss is as follows.
PartMixUp loss $\mathcal{L}_{\text{PM}}$ on a batch of embeddings $Z$ is computed by selecting the furthest positive $z_{pos}$ within the batch and the closest mixed up negative $z^{\prime}$ for each anchor $z_a$:
\begin{equation}
\label{ex:partmixup-loss}
    \mathcal{L}_{\text{PM}}(Z) = 
    \sum_{z_a \in Z}
    \Big[
          m + 
          \max_{z_{pos} \in Z} D\big(z_a, z_{pos}\big) - 
          \min_{z^{\prime} \in \widetilde{Z}} D\big(z_a, z^{\prime}\big)
    \Big]
\end{equation}
where $\widetilde{Z}$ is composed of $Z$ by replacing some semantic part embeddings for each anchor $z_a \in Z$ with corresponding part embeddings from another identity.

The number of part embeddings controls the difficulty of generated negative pairs shared in the pair: the more parts are shared, the harder the negative pair becomes (Figure~\ref{fig:partmixup}).
We will use the notation $\text{PM}(a)$ for the PartMixUp loss with the maximum number of shared part embeddings equal to $a$. 
Replacing parts is performed between part embeddings and not in the image pixels. 
Cutting and pasting image parts is prone to errors in a part detector.

The key difference between our PartMixUp loss and the previous variations of the triplet loss \cite{batch-hard-triplet, Song2016DeepML} is that it minimizes the distances for negative pairs by creating new samples rather than searching for them within the batch.
The benefit of PartMixUp loss is the improved discriminative ability of semantic part embeddings.
In the ablation study (Sec.~\ref{sec:experiments-ablation}), we show that adding PartMixUp loss improves the model performance.

\textbf{Total loss.}
In addition to the triplet and PartMixUp losses, the model is optimized with a cross-entropy loss widely used for person re-identification \citep{bag-tricks-person-reid, Zhai2019InDO}. 
A classification layer is added at the beginning of each pseudo-labeling iteration with the number of outputs being the number of known training identities (labeled and pseudo-labeled).
The overall objective is a weighted sum of all loss functions defined as:

\begin{equation}
\label{eq:total-loss}
    \mathcal{L} = 
        \lambda_{\text{CE}} \mathcal{L}_{\text{CE}} + 
        \lambda_{\text{T}} \mathcal{L}_{\text{T}} + 
        \lambda_{\text{PM}} \mathcal{L}_{\text{PM}}
\end{equation}
where $\lambda_{\text{CE}}, \lambda_{\text{T}}$ and $\lambda_{\text{PM}}$ are the weighting factors for each loss.               
\section{Experiments}
\label{sec:experiments}

\subsection{Datasets}
\label{sec:experiments-datasets}

We evaluate our method on three benchmarks in person re-identification: Market-1501~\citep{market-dataset}, DukeMTMC-reID~\citep{duke-dataset} and CUHK03~\citep{cuhk03-dataset}.

\textbf{Market-1501} \citep{market-dataset} contains 32,668 images for 1501 identities captured from 6 cameras placed in front of a campus supermarket.
The standard evaluation protocol \citep{market-dataset} splits the dataset into fixed subsets: 12,936 images of 751 identities in the training subset, and 15,913 images in the gallery subset and 3,368 images in the query subset of 750 test identities (disjoint with training identities).
During testing, query images are used to retrieve matching images in the gallery set.
The bounding boxes are computed using Deformable Part Model (DPM) \citep{Felzenszwalb2009ObjectDW} making it close to realistic settings.

\textbf{DukeMTMC-reID} \citep{duke-dataset} contains 36,411 images for 1,404 identities captured from 8 cameras, which is a subset of the pedestrian tracking dataset. The dataset is split into three fixed subsets: 702 identities with 16,522 images are used for training, and 2,228 images from other 702 identities are used for query images retrieving the rest 17,661 gallery images.
The semi-supervised settings for DukeMTMC-reID are the same as for Market-1501 dataset.

\textbf{CUHK03} dataset \citep{cuhk03-dataset} contains 14,097 images of 1,467 identities. 
We use the first evaluation protocol \citep{cuhk03-dataset} (like the majority of existing works) that splits the dataset randomly 20 times, and the gallery for testing has 100 identities each time. 
We evaluate on bounding boxes automatically detected by DPM \citep{Felzenszwalb2009ObjectDW}.

\subsection{Evaluation protocol}
\label{sec:experiments-evaluation}

We follow the semi-supervised setting of recent works \citep{SSL-Liu2020SemanticsGuidedCW, SSL-person-reid-multi-view} where only a 1/3 of identities is labeled and identities of labeled and unlabeled subsets are disjoint.

We assess the performance with two evaluation metrics: Cumulated Matching Characteristics (CMC) which treats the re-identification task as a ranking problem, and mean Average Precision (mAP) which treats it as an object retrieval problem. 
Similar to previous works \citep{SSL-person-reid-multi-view, SSL-Xin2019DeepSL, SSL-Chang2020TransductiveSM, SSL-Pan2019MultiViewAM, SSL-Liu2020SemanticsGuidedCW}, on Market-1501 and on DukeMCMT-reID, we report CMC at rank-1 and mAP in a single-query mode. 
The performance on CUHK03 is evaluated at rank-1, rank-5, rank-10 and rank-20 in a single-shot mode.

\subsection{Implementation details}
\label{sec:experiments-implementation}

We use ResNet50 \citep{resnet} architecture as a backbone like most other semi-supervised re-id. 
We build on the part-based convolutional baseline (PCB) proposed in \citep{pcb-model} that outputs part-based embeddings by pooling feature maps computed by the backbone network over regions of interest (ROI).
PCB works with a coarse part detection which is a split into six equal horizontal stripes.
We use PCB without refined part pooling (RPP) proposed in the same work.
Utilizing a more sophisticated part detection method may further improve the results.

We perform five pseudo-labeling iterations with 100 epochs in each iteration.
The model is trained with Adam optimizer \cite{adam}. 
The initial learning rate is set to 0.001 with decay at 60th and 80th epochs in each iteration. The decay factor is set as 0.1.
Furthermore, input images are re-sized to 128$\times$384 as for the original PCB model. For data augmentation, we randomly translate and horizontally flip the images.
For the PartMixUp loss, at most five out of six parts are replaced.

Each training batch contains 10 labeled identities and 10 pseudo-labeled identities with 6 images per identity.
At the first pseudo-labeling iteration, when the data has not been pseudo-labeled yet, the training batch contains only labeled data.
Our method is implemented with PyTorch \citep{pytorch} and TorchReId \citep{Zhou2019TorchreidAL}.
We train the model on one Tesla M40 GPU 12GB.

\subsection{Semi-supervised evaluation}
\label{sec:experiments-results-ss-same-domain}

\begin{table}
\caption{Performance comparison of our method with other methods on different datasets with 1/3 labeled data. BIL - the baseline image-level embeddings model, PB - the part-based model, PM - the PartMixUp loss.}
\label{tab:results-ours-all-datsets}
\begin{tabular}{l|cccccc}
\multirow{2}{*}{Method} & \multicolumn{2}{c}{Market-1501} & \multicolumn{2}{c}{DukeMTMC} & \multicolumn{2}{c}{CUHK03} \\
 & \multicolumn{1}{c}{Rank-1} & \multicolumn{1}{c}{mAP} & \multicolumn{1}{c}{Rank-1} & \multicolumn{1}{c}{mAP} & \multicolumn{1}{c}{Rank-1} & \multicolumn{1}{c}{Rank-5} \\ \hline
Supervised BIL & 75.1 & 53.3 & 65.0 & 49.4 & 48.3 & 73.6 \\
Semi-supervised BIL & 79.5 & 62.1 & 69.4 & 50.2 & 53.8 & 76.8 \\
Semi-supervised PB & 87.1 & 69.7 & 75.0 & 58.4 & 56.7 & 79.6 \\
Ours PB & 90.7 & 76.2 & 81.1 & 66.4 & 58.4 & 82.5 \\
Ours PB+PM & \textbf{91.5} & \textbf{76.7} & \textbf{82.4} & \textbf{67.5} & \textbf{60.1}  & \textbf{84.7}
\end{tabular}
\end{table}

Table~\ref{tab:results-ours-all-datsets} shows results with different methods on Market-1501, DukeMTMC-reID and CUHK03 datasets with 1/3 of labeled data.
As a baseline, we train the backbone architecture with image-level embeddings (BIL) in a supervised manner on the labeled subset.
Semi-supervised BIL is trained with pseudo-labels computed using agglomerative clustering.
A semi-supervised part-based (PB) model is trained with pseudo-labels computed using concatenated part-based embeddings.
Our part-based (PB) model is trained with pseudo-labels computed using consensus-clustering of part-based embeddings.
Finally, the PB model with PartMixUp loss (PB+PM) is trained as the previous model with the additional PartMixUp loss component (Equation~\ref{eq:total-loss}).

The experiment results on three datasets (Table~\ref{tab:results-ours-all-datsets}) demonstrate that consensus-clustering of part-based embeddings (our PB) outperforms both clustering of image-level embeddings (semi-supervised BIL) and clustering of concatenated part-based embeddings (semi-supervised PB).
Moreover, learning part-based embeddings with PartMixUp loss (our PB+PM) further improves both Rank-1 and mAP metrics.

\begin{figure}
    \begin{subfigure}{.5\textwidth}
      \centering
      \includegraphics[width=\linewidth]{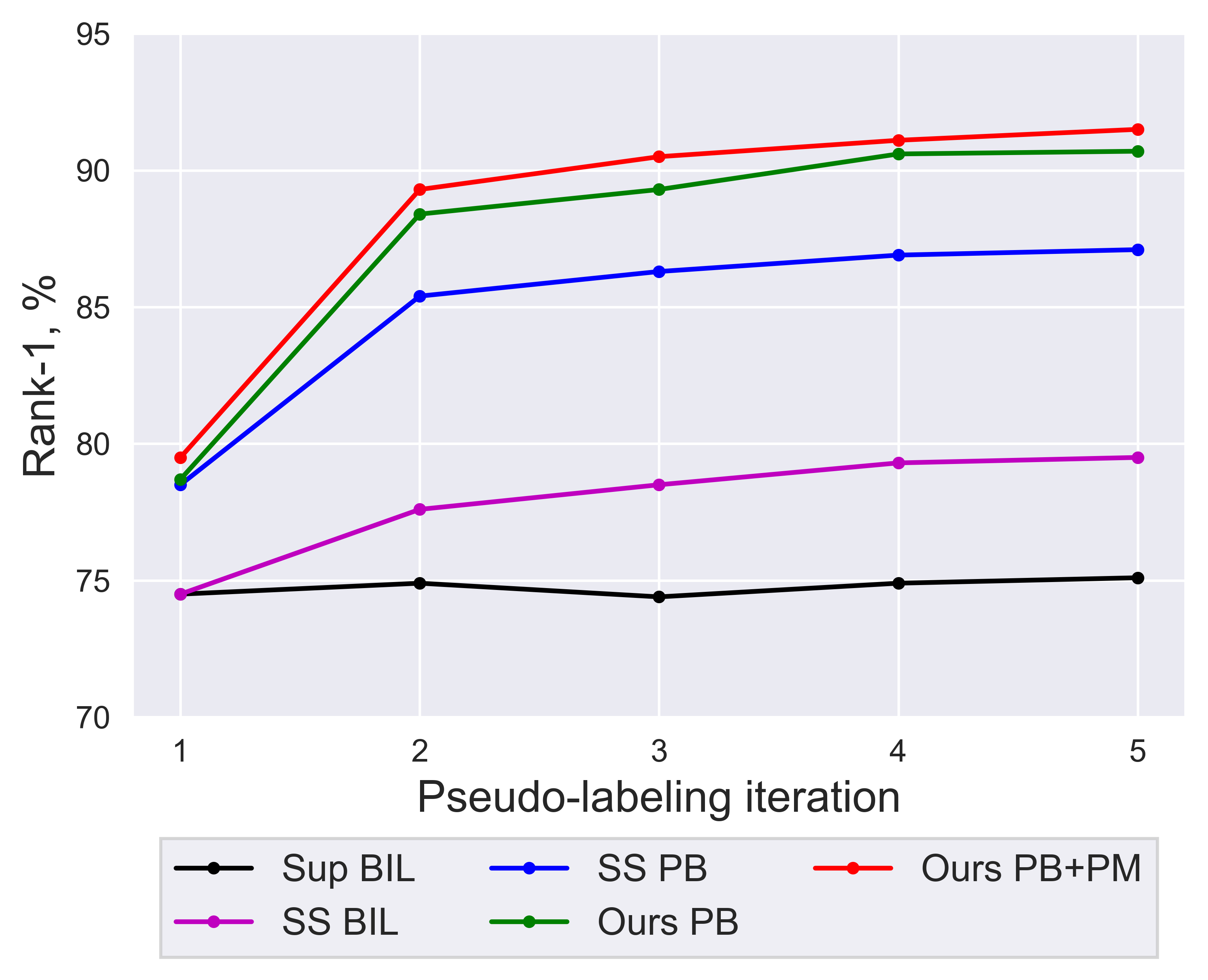}
      \caption{Rank-1.}
      \label{fig:rank1-progress-iterations}
    \end{subfigure}%
    \begin{subfigure}{.5\textwidth}
      \centering
      \includegraphics[width=\linewidth]{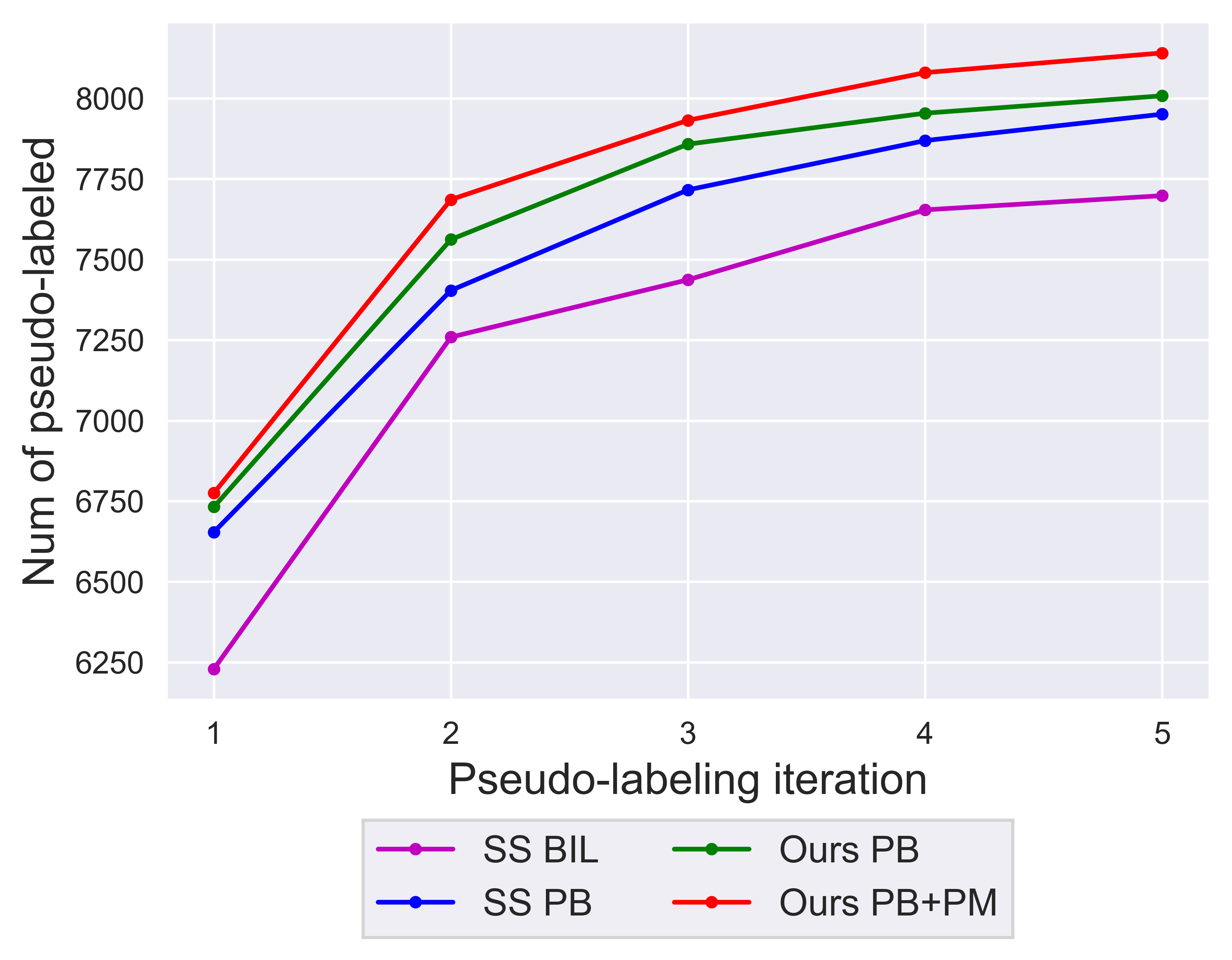}
      \caption{Number of pseudo-labeled images.}
      \label{fig:numpl-progress-iterations}
    \end{subfigure}
    \caption{Progress of methods over pseudo-labeling iterations on Market-1501 dataset with 1/3 labeled data. BIL - backbone network with image-level embeddings, PB - part-based model and PM - PartMixUp loss. Sup and SS - supervised and semi-supervised settings.}
    \label{fig:progress-iterations}
\end{figure}

Figures~\ref{fig:rank1-progress-iterations} and \ref{fig:numpl-progress-iterations} show the progress of Rank-1 and the number of pseudo-labeled images for different methods over pseudo-labeling iterations.
Supervised training of BIL model does not utilize pseudo-labeled images so it is not shown in Figure~\ref{fig:numpl-progress-iterations}.
A sharp increase in Rank-1 is observed in the second iteration due to the fact that a bulk of pseudo-labels is added to the training after the model has been pretrained on the labeled data in the first iteration.

\subsection{Comparison with state-of-the-art methods}
\label{sec:experiments-results-vs-sota}

We compare the performance of our method with existing semi-supervised person re-id which use similar experimental settings on Market-1501 and  DukeMTMC-reID (Table~\ref{tab:results-sota-market-duke}) and CUHK03 (Table~\ref{tab:results-sota-cuhk}).
We select the setting with 1/3 labeled data that most existing works use to report the results.
Our approach outperforms previous methods at both CMC and mAP metrics.
Our method achieves 91.5\% in Rank-1 and 76.7\% in mAP on Market-1501 without the assumption on the number of identities in unlabeled data.
Tables~\ref{tab:results-sota-market-duke} and \ref{tab:results-sota-cuhk} show that the performance of our method on 1/3 labeled data is close to the performance of the same part-based model \citep{pcb-model} on the whole labeled dataset, e.g. Rank-1 82.4\%  of our method versus 82.6\% with full supervision on DukeMTMC-reID and Rank-1 91.5\%  of our method versus 92.3\% with full supervision on Market-1501.

\begin{table}
    \caption{Performance comparison of our method with other semi-supervised re-id on Market-1501 and DukeMTMC-reID (Results in \%). RN50 - ResNet50 \cite{resnet}, DN121 - DenseNet121 \cite{Huang2017DenselyCC}, Wrn50 \cite{Zagoruyko2016WideRN}.}
    \label{tab:results-sota-market-duke}
    \centering
    \setlength\tabcolsep{3pt}
    \renewcommand{\arraystretch}{1.15}
    \begin{tabular}{l|l|ccc|ccc}
        \multicolumn{1}{c|}{\multirow{2}{*}{Method}} & \multicolumn{1}{c|}{\multirow{2}{*}{Backbone}} & \multicolumn{3}{c}{Market-1501} & \multicolumn{3}{c}{DukeMTMC-reID} \\ 
        \multicolumn{1}{c|}{} & \multicolumn{1}{c|}{} & Rank-1 & Rank-5 & mAP & Rank-1 & Rank-5 & mAP \\ 
        \hline
        \multicolumn{6}{l}{\textit{1/3 labeled data:}}
        \\
        MVC \citep{SSL-person-reid-multi-view} & RN50 & 72.2 & - & 49.6 & 52.9 & - & 33.6 \\
        MVC \citep{SSL-person-reid-multi-view} & RN50+Wrn50 & 75.2 & - & 52.6 & 55.7 & - & 37.8 \\
        MVSPC \citep{SSL-Xin2019DeepSL} & RN50 & 71.5 & 86.2 & 53.2 & 58.5 & 73.7 & 37.4 \\
        MVSPC \citep{SSL-Xin2019DeepSL} & RN50+DN121 & 80.1 & 91.9 & 62.8 & 70.8 & 82.2 & 50.3 \\
        TSSML \cite{SSL-Chang2020TransductiveSM} & RN50 & 86.4 & 95.0 & 69.1 & 72.7 & 85.2 & 53.2 \\
        MVMIC \citep{SSL-Pan2019MultiViewAM} & RN50+DN121 & 88.9 & 95.8 & 73.2 & 81.8 & 90.9 & 66.2 \\
        \hline
        Ours & RN50 & \textbf{91.5} & \textbf{96.5} & \textbf{76.7} & \textbf{82.4} & \textbf{91.3} & \textbf{67.5} \\
        \multicolumn{6}{l}{\textit{Fully supervised method:}}
        \\
        PCB \citep{pcb-model} & RN50 & 92.3  & 97.2 & 77.4 & 82.6 & - & 68.8
    \end{tabular}
\end{table}

\begin{table}
    \caption{Performance comparison of our method with other semi-supervised re-id on CUHK03 dataset (\%). RN50 - ResNet50 \cite{resnet}, Wrn50 - WideResnet50 \citep{Zagoruyko2016WideRN}.}
    \label{tab:results-sota-cuhk}
    \centering
    \renewcommand{\arraystretch}{1.15}
    \begin{tabular}{l|l|cccc}
        \multicolumn{1}{c|}{\multirow{2}{*}{Method}} & \multicolumn{1}{c|}{\multirow{2}{*}{Backbone}} & 
        \multicolumn{4}{c}{CUHK03} \\ 
        \multicolumn{1}{c|}{} & \multicolumn{1}{c|}{} & 
        Rank-1 & Rank-5 & Rank-10 & Rank-20\\ 
        \hline
        \multicolumn{6}{l}{\textit{1/3 labeled data:}}
        \\
        MVC \citep{SSL-person-reid-multi-view} & RN50 &  50.6 & 78.2 & 88.3 & 95.3 \\
        MVC \citep{SSL-person-reid-multi-view} & RN50+Wrn50 &  53.2 & 80.1 & 88.9 & 95.6 \\
        \hline
        Ours & RN50 &  \textbf{60.1} & \textbf{84.7} & \textbf{93.5} & \textbf{98.2} \\
        \multicolumn{6}{l}{\textit{Fully supervised method:}}
        \\
        PCB \citep{pcb-model} & RN50 & 61.3  & - & - & -
        
    \end{tabular}
\end{table}

\subsection{Ablation study}
\label{sec:experiments-ablation}

\begin{figure}
  \centering
  \begin{minipage}[b]{0.49\textwidth}
    \centering
    \includegraphics[width=\textwidth]{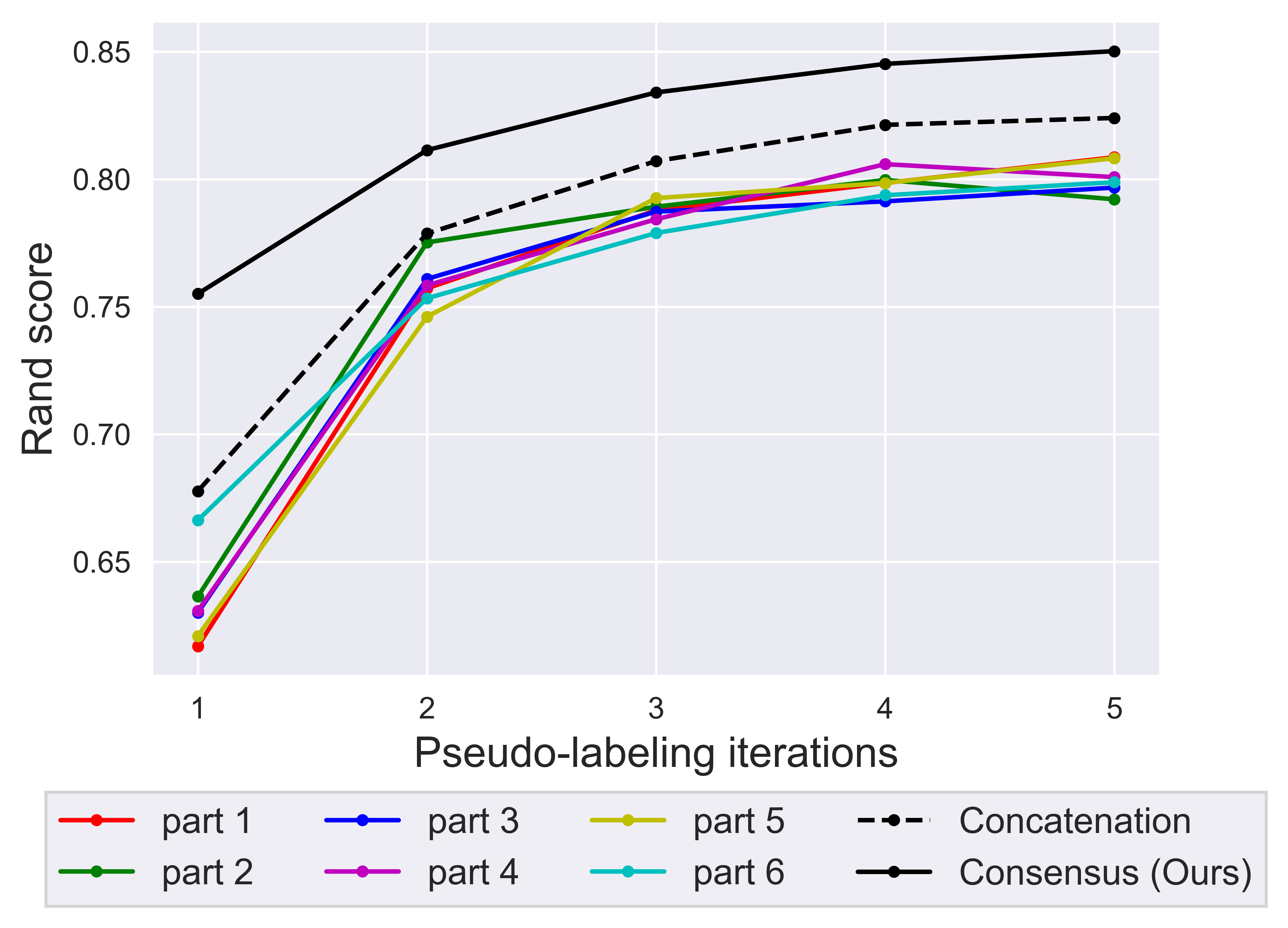}
    \caption{Rand score (the higher the better) of consensus clustering versus clustering image parts individually. Results are on Market-1501 with 1/3 labeled data.}
    \label{fig:clustering-rand-score}
  \end{minipage}
  \hfill
  \begin{minipage}[b]{0.49\textwidth}
    \centering
    \includegraphics[width=\textwidth]{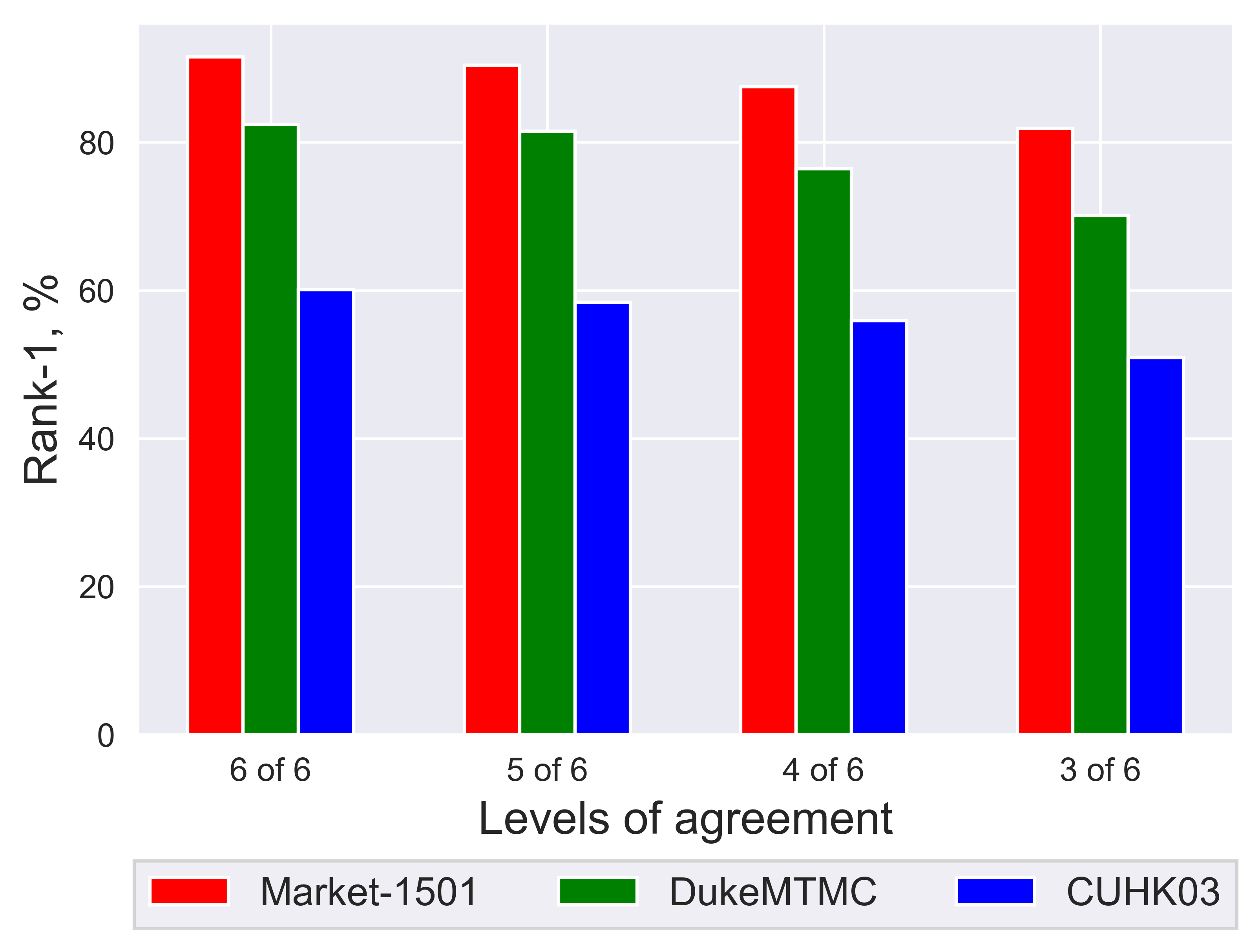}
    \caption{Rank-1 of semi-supervised re-id with levels of agreement from 3 to 6 parts in consensus clustering. Results are on different datasets with 1/3 labeled data.}
    \label{fig:consensus-agreement}
  \end{minipage}
\end{figure}

In this section, we review and evaluate various components and design choices in the proposed method.

\textbf{The effect of consensus clustering.}
To evaluate the importance of consensus clustering, we analyze the Rand index \cite{Hubert1985ComparingP} of cluster assignment on the unlabeled subset using available ground truth labels (used only for evaluation).
The Rand Index is a similarity measure between two cluster partitions computed by considering the ratio of pairs that are assigned to the same clusters in the predicted and true assignments.
Figure~\ref{fig:clustering-rand-score} shows the Rand Index of clustering for each semantic part, concatenation of part embeddings and our consensus clustering.
We observe that the highest Rand Index of 0.85 is achieved with our consensus clustering comparing to the concatenation of parts (the Rand Index 0.82) and clustering by each semantic part separately (the Rand Index 0.79 - 0.80).

\textbf{The influence of agreement in consensus clustering.}
In Figure~\ref{fig:consensus-agreement}, we show the results of experiments with varying levels of agreement in consensus clustering on different datasets.
We vary the number of semantic parts required for agreement in consensus clustering and perform experiments with strict 100\% agreement (6 out of 6 parts), 83\% agreement (5 out of 6 parts), 66\% (4 out of 6 parts) and 50\% (3 out of 6 parts).
The best results are achieved when all 6 out of 6 parts agree on the cluster assignment.
The performance decreases slightly when 5 out of 6 parts agree, from 91.5\% to 90.4\% on 1/3 labeled data on the Market-1501 dataset.
The performance degrades significantly with the agreement between only 4 or 3 parts (87.5\% and 81.9\% respectively).
Experiments confirm that strict agreement in consensus clustering is essential in our semi-supervised method.

\textbf{The influence of the clustering method.}
We evaluate our method with two other clustering algorithms that do not require the pre-defined number of clusters, Affinity Propagation \citep{AP} and DBSCAN \citep{DBSCAN}.
Experiments are conducted with default parameters for the clustering algorithms on the Market-1501 dataset with 1/3 and 1/6 labeled data.
Table~\ref{tab:ablation-clustering} shows the number of detected clusters versus the number of ground truth identities at the first pseudo-labeling iteration.
Rank-1 and mAP are compared after the first iteration.
We observe that DBSCAN failed to identify any clusters (similar results are observed in \citep{Qi2020ProgressiveCS}) so DBSCAN requires tuning parameters based on heuristics that we try to avoid.
Agglomerative clustering identifies more clusters than Affinity Propagation at the first iteration and yields better performance with the assigned pseudo-labels (rank-1 88.1\% versus 87.4\% and mAP 72.6\% versus 70.3\%).

\begin{table}
    \caption{The influence of clustering algorithm on Market-1501 dataset with 1/3 labeled data. The results are reported after one pseudo-labeling iteration.}
    \label{tab:ablation-clustering}
    \centering
    \setlength\tabcolsep{3pt}
    \begin{tabular}{l|cccc|cccc}
    \multirow{2}{*}{Method} & \multicolumn{4}{c|}{1/3 labeled} & \multicolumn{4}{c}{1/6 labeled} \\
     & \begin{tabular}[c]{@{}c@{}}Num\\ Clusters\end{tabular} & \begin{tabular}[c]{@{}c@{}}Num\\ Id\end{tabular} & Rank-1 & mAP & \begin{tabular}[c]{@{}c@{}}Num\\ Clusters\end{tabular} & \begin{tabular}[c]{@{}c@{}}Num\\ Id\end{tabular} & Rank-1 & mAP \\ \hline
    Affinity Prop. \cite{AP} & 443 & 501 & 87.4 & 70.3 & 495 & 626 & 81.3 & 59.9 \\
    DBSCAN \cite{DBSCAN} & 1 & 501 & - & - & 1 & 626 & - & - \\
    Agglomerative \cite{szekely2005hierarchical} & 461 & 501 & 88.1 & 72.6 & 554 & 626 & 84.8 & 65.1
    \end{tabular}
\end{table}

\textbf{The influence of PartMixUp loss.}
We observe that PartMixUp loss improves the re-id performance of the part-based PCB model \citep{pcb-model} in the supervised setting and our method in the semi-supervised setting (Table~\ref{tab:ablation-pm-loss}).

\begin{table}
    \caption{The influence of PartMixUp loss on performance of PCB model in the fully supervised setting and our method (with PCB model in its core) in the semi-supervised setting.}
    \label{tab:ablation-pm-loss}
    \centering
    \setlength\tabcolsep{3pt}
    \renewcommand{\arraystretch}{1.15}
    \begin{tabular}{l|c|ccc|ccc}
        \multicolumn{1}{c|}{\multirow{2}{*}{Method}} & \multicolumn{1}{c|}{\multirow{2}{*}{PM Loss}} & \multicolumn{3}{c}{Market-1501} & \multicolumn{3}{c}{DukeMTMC-reID} \\ 
        \multicolumn{1}{c|}{} & \multicolumn{1}{c|}{} & Rank-1 & Rank-5 & mAP & Rank-1 & Rank-5 & mAP \\ 
        \hline
        \multicolumn{6}{l}{\textit{Fully supervised method:}}
        \\
        PCB \citep{pcb-model} & \xmark & 92.3  & 97.2 & 77.4 & 82.6 & - & 68.8
        \\
        PCB \citep{pcb-model} & \cmark & \textbf{93.1}  & \textbf{97.4} & \textbf{78.2} & \textbf{83.2} & - & \textbf{69.8}
        \\
        \hline
        \multicolumn{6}{l}{\textit{1/3 labeled data:}}
        \\
        Ours & \xmark & 90.7 & 95.9 & 76.2 & 81.1 & 91.1 & 66.4  
        \\
        Ours & \cmark & \textbf{91.5} & \textbf{96.5} & \textbf{76.7} & \textbf{82.4} & \textbf{91.3} & \textbf{67.5} 
        \\
    \end{tabular}
\end{table}

\section{Conclusion}
In this paper, we propose a novel semi-supervised method for person re-id by consensus clustering of part-based embeddings.
Our method assigns pseudo-labels without any assumption about the number of identities in the unlabeled subset.
The pseudo-labels are assigned based on consensus clustering of part-based embeddings which yields better pseudo-labels than clustering of image-level embeddings or features from multiple CNNs.
The developed PartMixUp loss improves the discriminative ability of part-based features and further increases the performance of the model.
Our method utilizes only one CNN and is compatible with any CNN architecture as a backbone.
Extensive experiments in various settings on multiple person re-id datasets confirm the effectiveness of the proposed approach.

\section*{Acknowledgements}
We acknowledge continued support from the Queensland University of Technology (QUT) through the Centre for Robotics.
The research is supported by Research Training Program (RTP) scholarship from Australian government.
Computational resources and services used in this work were provided by the HPC and Research Support Group, Queensland University of Technology, Brisbane, Australia.


\bibliographystyle{elsarticle-num}
\bibliography{references}

\begin{thebibliography}{10}
\expandafter\ifx\csname url\endcsname\relax
  \def\url#1{\texttt{#1}}\fi
\expandafter\ifx\csname urlprefix\endcsname\relax\def\urlprefix{URL }\fi
\expandafter\ifx\csname href\endcsname\relax
  \def\href#1#2{#2} \def\path#1{#1}\fi

\bibitem{market-dataset}
L.~Zheng, L.~Shen, L.~Tian, S.~Wang, J.~Wang, Q.~Tian, Scalable person
  re-identification: A benchmark, in: Proc. ICCV, 2015.

\bibitem{duke-dataset}
E.~Ristani, F.~Solera, R.~S. Zou, R.~Cucchiara, C.~Tomasi, Performance measures
  and a data set for multi-target, multi-camera tracking, in: ECCV Workshops,
  2016.

\bibitem{part-loss}
H.~Yao, S.~Zhang, R.~Hong, Y.~Zhang, C.~Xu, Q.~Tian, Deep representation
  learning with part loss for person re-identification, IEEE Transactions on
  Image Processing 28 (2019) 2860--2871.

\bibitem{Zhang2019DenselySA}
Z.~Zhang, C.~Lan, W.~Zeng, Z.~Chen, Densely semantically aligned person
  re-identification, in: Proc. CVPR, 2019.

\bibitem{batch-hard-triplet}
A.~Hermans, L.~Beyer, B.~Leibe, {In Defense of the Triplet Loss for Person
  Re-Identification}, arXiv:1703.07737 (2017).

\bibitem{bag-tricks-person-reid}
H.~Luo, Y.~Gu, X.~Liao, S.~Lai, W.~Jiang, Bag of tricks and a strong baseline
  for deep person re-identification, in: Proc. CVPRW, 2019.

\bibitem{Ghorbel2020FusingLA}
M.~Ghorbel, S.~Ammar, Y.~Kessentini, M.~Jmaiel, A.~Chaari, Fusing local and
  global features for person re-identification using multi-stream deep neural
  networks, Pattern Recognition and Artificial Intelligence 1322 (2020) 73 --
  85.

\bibitem{Iscen2019LabelPF}
A.~Iscen, G.~Tolias, Y.~Avrithis, O.~Chum, Label propagation for deep
  semi-supervised learning, in: Proc. CVPR, 2019.

\bibitem{SSL-person-reid-multi-view}
X.~Xin, J.~Wang, R.~Xie, S.~Zhou, W.~Huang, N.~Zheng, Semi-supervised person
  re-identification using multi-view clustering, Pattern Recognition 88 (2019)
  285--297.

\bibitem{SSL-Xin2019DeepSL}
Deep self-paced learning for semi-supervised person re-identification using
  multi-view self-paced clustering, in: Xiaomeng Xin and Xindi Wu and Yuechen
  Wang and Jinjun Wang, 2019.

\bibitem{SSL-Pan2019MultiViewAM}
S.~Pan, Y.~Wang, Y.~Chong, Multi-view and multi-information clustering for
  semi-supervised person re-identification, in: Proc. EEI, 2019, pp. 200--205.

\bibitem{Sun2018BeyondPM}
Y.~Sun, L.~Zheng, Y.~Yang, Q.~Tian, S.~Wang, Beyond part models: Person
  retrieval with refined part pooling, in: Proc. ECCV, 2018.

\bibitem{pcb-model}
Y.~Sun, L.~Zheng, Y.~Li, Y.~Yang, Q.~Tian, S.~Wang, Learning part-based
  convolutional features for person re-identification., IEEE Transactions on
  Pattern Analysis and Machine Intelligence (2019).

\bibitem{triplet-loss}
J.~Wang, Y.~Song, T.~Leung, C.~Rosenberg, J.~Wang, J.~Philbin, B.~Chen, Y.~Wu,
  Learning fine-grained image similarity with deep ranking, in: Proc. CVPR,
  2014.

\bibitem{Chen2017BeyondTL}
W.~Chen, X.~Chen, J.~Zhang, K.~Huang, Beyond triplet loss: A deep quadruplet
  network for person re-identification, in: Proc. CVPR, 2017.

\bibitem{Zhai2019InDO}
Y.~Zhai, X.~Guo, Y.~Lu, H.~Li, In defense of the classification loss for person
  re-identification, in: Proc. CVPRW, 2019.

\bibitem{Ding2019FeatureAP}
G.~Ding, S.~Zhang, S.~Khan, Z.~Tang, J.~Zhang, F.~Porikli, Feature
  affinity-based pseudo labeling for semi-supervised person re-identification,
  IEEE Transactions on Multimedia 21 (2019) 2891--2902.

\bibitem{Zhang2020SemisupervisedPR}
X.~Zhang, X.-Y. Jing, X.~Zhu, F.~Ma, Semi-supervised person re-identification
  by similarity-embedded cycle gans, Neural Computing and Applications (2020)
  1--10.

\bibitem{Ge2020SelfpacedCL}
Y.~Ge, D.~peng Chen, F.~Zhu, R.~Zhao, H.~Li, Self-paced contrastive learning
  with hybrid memory for domain adaptive object re-id, in: Proc. NeurIPS, 2020.

\bibitem{Fan2018UnsupervisedPR}
H.~Fan, L.~Zheng, Y.~Yang, Unsupervised person re-identification, ACM
  Transactions on Multimedia Computing, Communications, and Applications (TOMM)
  14 (2018) 1 -- 18.

\bibitem{SSL-Chang2020TransductiveSM}
X.~Chang, Z.~Ma, X.~Wei, X.~Hong, Y.~Gong, Transductive semi-supervised metric
  learning for person re-identification, Pattern Recognition 108 (2020) 107569.

\bibitem{SSL-Liu2020SemanticsGuidedCW}
C.~Liu, Y.-J. Li, S.~Chien, Y.-C.~F. Wang, Semantics-guided clustering with
  deep progressive learning for semi-supervised person re-identification,
  arXiv:2010.01148 (2020).

\bibitem{Qi2020ProgressiveCS}
L.~Qi, L.~Wang, J.~Huo, Y.~Shi, Y.~Gao, Progressive cross-camera soft-label
  learning for semi-supervised person re-identification, IEEE Transactions on
  Circuits and Systems for Video Technology 30 (2020) 2815--2829.

\bibitem{Wang2021GraphInducedCL}
M.~Wang, B.~Lai, J.~Huang, X.~Gong, X.~Hua, Graph-induced contrastive learning
  for intra-camera supervised person re-identification, IEEE Access 9 (2021)
  20850--20860.

\bibitem{Zhu2019IntraCameraSP}
X.~Zhu, X.~Zhu, M.~Li, V.~Murino, S.~Gong, Intra-camera supervised person
  re-identification: A new benchmark, in: Proc. ICCVW, 2019.

\bibitem{Li2018SemisupervisedRM}
J.~Li, A.~J. Ma, P.~Yuen, Semi-supervised region metric learning for person
  re-identification, International Journal of Computer Vision 126 (2018)
  855--874.

\bibitem{Bk2017OneShotML}
S.~Bak, P.~Carr, One-shot metric learning for person re-identification, in:
  Proc. CVPR, 2017.

\bibitem{Lin2020UnsupervisedPR}
Y.~Lin, L.~Xie, Y.~Wu, C.~Yan, Q.~Tian, Unsupervised person re-identification
  via softened similarity learning, 2020.

\bibitem{Ye2021DeepLF}
M.~Ye, J.~Shen, G.~Lin, T.~Xiang, L.~Shao, S.~Hoi, Deep learning for person
  re-identification: A survey and outlook, IEEE Transactions on Pattern
  Analysis and Machine Intelligence (2021).

\bibitem{Leng2020ASO}
Q.~Leng, M.~Ye, Q.~Tian, A survey of open-world person re-identification, IEEE
  Transactions on Circuits and Systems for Video Technology 30 (2020)
  1092--1108.

\bibitem{Yaghoubi2021SSSPRAS}
E.~Yaghoubi, A.~Kumar, H.~Proença, Sss-pr: A short survey of surveys in person
  re-identification, Pattern Recognition Letters 143 (2021) 50--57.

\bibitem{Lin2019ABC}
Y.~Lin, X.~Dong, L.~Zheng, Y.~Yan, Y.~Yang, A bottom-up clustering approach to
  unsupervised person re-identification, in: Proc. AAAI, 2019.

\bibitem{Yu2020UnsupervisedPR}
H.-X. Yu, A.~Wu, W.~Zheng, Unsupervised person re-identification by deep
  asymmetric metric embedding, Transactions on Pattern Analysis and Machine
  Intelligence 42 (2020) 956--973.

\bibitem{Maksai2017NonMarkovianGC}
A.~Maksai, X.~Wang, F.~Fleuret, P.~Fua, Non-markovian globally consistent
  multi-object tracking, in: Proc. ICCV, 2017.

\bibitem{Eom2019LearningDR}
C.~Eom, B.~Ham, Learning disentangled representation for robust person
  re-identification, in: Proc. NeurIPS, 2019.

\bibitem{Zheng2017UnlabeledSG}
Z.~Zheng, L.~Zheng, Y.~Yang, Unlabeled samples generated by gan improve the
  person re-identification baseline in vitro, in: Proc. ICCV, 2017.

\bibitem{Zheng2019JointDA}
Z.~Zheng, X.~Yang, Z.~Yu, L.~Zheng, Y.~Yang, J.~Kautz, Joint discriminative and
  generative learning for person re-identification, in: Proc. CVPR, 2019.

\bibitem{Deng2018ImageImageDA}
W.~Deng, L.~Zheng, G.~Kang, Y.~Yang, Q.~Ye, J.~Jiao, Image-image domain
  adaptation with preserved self-similarity and domain-dissimilarity for person
  re-identification, in: Proc. CVPR, 2018.

\bibitem{Yu2019UnsupervisedPR}
H.-X. Yu, W.~Zheng, A.~Wu, X.~Guo, S.~Gong, J.~Lai, Unsupervised person
  re-identification by soft multilabel learning, in: Proc. CVPR, 2019.

\bibitem{Zhong2019InvarianceME}
Z.~Zhong, L.~Zheng, Z.~Luo, S.~Li, Y.~Yang, Invariance matters: Exemplar memory
  for domain adaptive person re-identification, in: Proc. CVPR, 2019.

\bibitem{KAE-Net}
O.~Moskvyak, F.~Maire, F.~Dayoub, M.~Baktashmotlagh, Keypoint-aligned
  embeddings for image retrieval and re-identification, in: Proc. WACV, 2021.

\bibitem{imagenet}
O.~Russakovsky, J.~Deng, H.~Su, J.~Krause, S.~Satheesh, S.~Ma, Z.~Huang,
  A.~Karpathy, A.~Khosla, M.~S. Bernstein, A.~Berg, L.~Fei-Fei, Imagenet large
  scale visual recognition challenge, International Journal of Computer Vision
  115 (2015) 211--252.

\bibitem{Alqurashi2017ClusteringEM}
T.~Alqurashi, Clustering ensemble method, International Journal of Machine
  Learning and Cybernetics 10 (2017) 1227--1246.

\bibitem{szekely2005hierarchical}
G.~J. Szekely, M.~L. Rizzo, et~al., Hierarchical clustering via joint
  between-within distances: Extending ward's minimum variance method, Journal
  of classification 22~(2) (2005) 151--184.

\bibitem{ward1963hierarchical}
J.~H. Ward~Jr, Hierarchical grouping to optimize an objective function, Journal
  of the American Statistical Association 58~(301) (1963) 236--244.

\bibitem{AP}
B.~J. Frey, D.~Dueck, Clustering by passing messages between data points,
  Science 315~(5814) (2007) 972--976.

\bibitem{DBSCAN}
M.~Ester, H.-P. Kriegel, J.~Sander, X.~Xu, A density-based algorithm for
  discovering clusters in large spatial databases with noise, in: Proc. of
  Conference on Knowledge Discovery and Data Mining, 1996, p. 226–231.

\bibitem{VegaPons2011ASO}
S.~Vega-Pons, J.~Ruiz-Shulcloper, A survey of clustering ensemble algorithms,
  International Journal of Pattern Recognition and Artificial Intelligence 25
  (2011) 337--372.

\bibitem{Fred2005CombiningMC}
A.~Fred, A.~K. Jain, Combining multiple clusterings using evidence
  accumulation, IEEE Transactions on Pattern Analysis and Machine Intelligence
  27 (2005) 835--850.

\bibitem{Song2016DeepML}
H.~O. Song, Y.~Xiang, S.~Jegelka, S.~Savarese, Deep metric learning via lifted
  structured feature embedding, in: Proc. CVPR, 2016.

\bibitem{cuhk03-dataset}
W.~Li, R.~Zhao, T.~Xiao, X.~Wang, Deepreid: Deep filter pairing neural network
  for person re-identification, in: Proc. CVPR, 2014.

\bibitem{Felzenszwalb2009ObjectDW}
P.~F. Felzenszwalb, R.~B. Girshick, D.~A. McAllester, D.~Ramanan, Object
  detection with discriminatively trained part based models, IEEE Transactions
  on Pattern Analysis and Machine Intelligence 32 (2009) 1627--1645.

\bibitem{resnet}
K.~He, X.~Zhang, S.~Ren, J.~Sun, Deep residual learning for image recognition,
  in: Proc. CVPR, 2016.

\bibitem{adam}
D.~P. Kingma, J.~Ba, Adam: A method for stochastic optimization, in: Proc.
  ICLR, 2015.

\bibitem{pytorch}
A.~Paszke, S.~Gross, F.~Massa, A.~Lerer, J.~Bradbury, G.~Chanan, T.~Killeen,
  Z.~Lin, N.~Gimelshein, L.~Antiga, A.~Desmaison, A.~Kopf, E.~Yang, Z.~DeVito,
  M.~Raison, A.~Tejani, S.~Chilamkurthy, B.~Steiner, L.~Fang, J.~Bai,
  S.~Chintala, Pytorch: An imperative style, high-performance deep learning
  library 32 (2019) 8024--8035.

\bibitem{Zhou2019TorchreidAL}
K.~Zhou, T.~Xiang, Torchreid: A library for deep learning person
  re-identification in pytorch, arXiv:1910.10093 (2019).

\bibitem{Huang2017DenselyCC}
G.~Huang, Z.~Liu, K.~Q. Weinberger, Densely connected convolutional networks,
  in: Proc. CVPR, 2017.

\bibitem{Zagoruyko2016WideRN}
S.~Zagoruyko, N.~Komodakis, Wide residual networks, in: Proc. BMVC, 2016.

\bibitem{Hubert1985ComparingP}
L.~Hubert, P.~Arabie, Comparing partitions, Journal of Classification 2 (1985)
  193--218.

\end{thebibliography}



\end{document}